# Vacuum Driven Auxetic Switching Structure and Its Application on a Gripper and Quadruped

Shuai Liu*, Sheeraz Athar*, *Student Member, IEEE* and Michael Yu Wang, *Fellow, IEEE*

*Abstract*— The properties and applications of auxetics have been widely explored in the past years. Through proper utilization of auxetic structures, designs with unprecedented mechanical and structural behaviors can be produced. Taking advantage of this, we present the development of novel and low-cost 3D structures inspired by a simple auxetic unit. The core part, which we call the body in this paper, is a 3D realization of 2D rotating squares. This body structure was formed by joining four similar structures through softer material at the vertices. A monolithic structure of this kind is accomplished through a custom-built multi-material 3D printer. The model works in a way that, when torque is applied along the face of the rotational squares, they tend to bend at the vertex of the softer material, and due to the connected-ness of the design, a proper opening and closing motion is achieved. To demonstrate the potential of this part as an important component for robots, two applications are presented: a soft gripper and a crawling robot. Vacuum-driven actuators move both the applications. The proposed gripper combines the benefits of two types of grippers whose fingers are placed parallel and equally spaced to each other, in a single design. This gripper is adaptable to the size of the object and can grasp objects with large and small cross- sections alike. A novel bending actuator, which is made of soft material and bends in curvature when vacuumed, provides the grasping nature of the gripper. Crawling robots, in addition to their versatile nature, provide a better interaction with humans. The designed crawling robot employs negative pressure-driven actuators to highlight linear and turning locomotion.

## I. INTRODUCTION

Auxetics deform in an unusual manner and have many impressive properties. One of which is that these structures can have a negative Poisson's ratio [1], [2], [3]. When stretched, they become thicker perpendicular to the applied force. In material and structure levels, auxetics are supposed to have high-energy absorption capability, thus nowadays researchers are seeking to use this application in body armor, shock-absorbing structures, etc. [4].

Plenty of research has been conducted into auxetic Metamaterials. It is found that even the simplest geometries can achieve auxetic behavior [1], [5]. By connecting triangles or squares along vertices, the structure generated is auxetic. Different kinds of cell units can give various mechanical parameters.

As there are so many unique benefits from these kinds of design, we propose an application of one type of auxetics, called rotating squares. We take advantage of its successive

All authors are with the Department of Mechanical and Aerospace Engineering, The Hong Kong University of Science and Technology. sliubw@connect.ust.hk, sathar@connect.ust.hk, mywang@ust.hk
*The authors have equally contributed to this work

deformation and combine that with a flexible gripper and quadruped robot design. This structure adds two more degrees of freedom (DOFs) to the gripper and crawling robot, while keeping the structure stable. These added DOFs can be used to change the working mode of the gripper and crawling robot.

Most flexible grippers can be divided into two categories, first are those grippers that have fingers or tentacles oriented opposite to each other [6], called the parallel gripper. The second category consists of grippers in which actuators that are equally spaced around its geometric center point [7], [8], [9], called the cross-link gripper in this paper. These kinds of designs aim to deal with most common grasping cases. However, in the practical application, both equally and opposite spaced actuators are always necessary. To tackle this issue, many widely used mechanical robotic hands use servos to increase their mobility, thereby increasing complexity. Compared to these traditional robotic hands, the gripper we are proposing overcomes this problem by combining both modes in a single design. Moreover, our design is much lower in cost, and has flexibility and adaptability as good as pure soft grippers because of the soft components we applied.

Recently, Tang presented programmable Kirigami structures that can control the direction of a soft quadruped crawling robot [10]. The Kirigami sheets presented in his paper also have auxetic properties. However, the structure proposed in the paper has more living hinges than simple rigid auxetic units, which can decrease the structural stiffness and make controlling the deformation of the whole structure extremely difficult. In addition, the turning structure has no locking design, so it needs a continuous supply of air to keep its configuration. This paper presents a design that addresses this problem by using magnets as a locking mechanism.

Overall, the design we are proposing is very efficient as it combines the benefit of structural functionalities with robotics and manipulation. Moreover, the second application of the design i.e. a crawling robot also provides numerous advantages as crawling robots interact better with humans, are less affected by their environment and give a multi-faceted application ranging from search and rescue to drug delivery [11], [12], [13], [14].

The contributions of our work include the development of low-cost auxetic turning structures and its use in a novel gripper that has the qualities of both parallel and cross-link grasping capabilities. In addition, we also designed a crawling robot for in-plane navigation. The main innovation of this study is the development of a modularized vacuum rotating actuator, which actuates both the above-mentioned applications.

## II. DESIGN AND FABRICATION OF THE AUXETIC BODY AND ACTUATORS

### A. Auxetic Body

In our design, the auxetic body is composed of four connected pentagons, as shown in Fig. 1. Rotating pentagons are a simple extension of rotating squares; they both have bi-stabilities. The only difference is that rotating pentagons have less moving range because of the body contact.

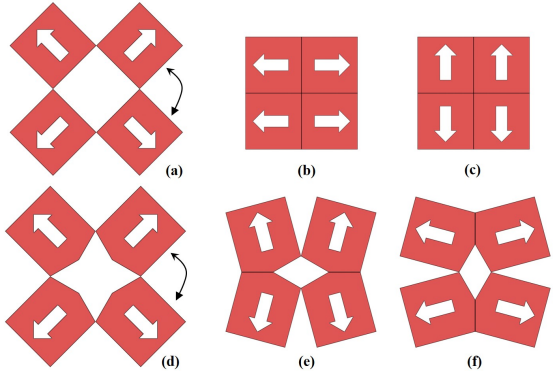

Fig. 1. (a)-(c) Three stable states of rotating squares; (d)-(f) Three stable states of rotating pentagons

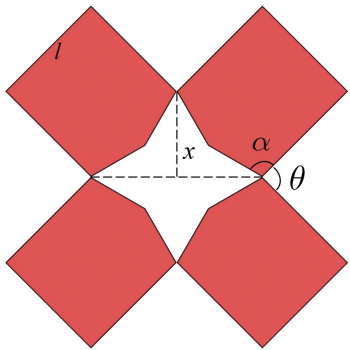

Fig. 2. Geometric definitions of rotating pentagons

The reason we gave up rotating squares is that, when the structure was fully folded, all the squares became closely packed to each other, leaving no room for the placement of an actuator. Therefore, to mount an actuator we modified the structure to rotating pentagons. Moreover, the angle between two adjacent edges in rotating squares is 180°, which means if the structure is stabilized at some folded state as in Fig. 1(b) or (c), it will be extremely hard for it to get rid of that state. The force that pneumatic actuators can apply on the surfaces of squares will be largely reduced because of the parallel surfaces, which reduces the force arms around the rotating center.

The angle between two adjacent edges is $\alpha$, the angle that adjacent units have is $\theta$, the width of the units is $l$. It could easily derive that $x = l\sin(\theta/2)$. We can see that the moving range of $\theta$ for the unit is from 0 to 360°-2$\alpha$.

### B. Vacuum Actuators

To actuate the auxetics, we constructed a novel vacuum rotating actuator. Shown in Fig. 3, the shape of the actuator is inspired by origami structures [15]. This design ensures the actuator has a large deformation ratio. However, if we simply deploy origami shapes on the chamber, it will not operate effectively because application of the vacuum will cause it to slacken, collapse, or distort, and it will not be able to maintain the shape. Therefore, we added inner support structures, as shown in Fig. 3(b), to avoid collapsing. These supports realized the expected functionalities. The actuator can be folded exactly along the hinges as shown in Fig. 3(e)-(g).

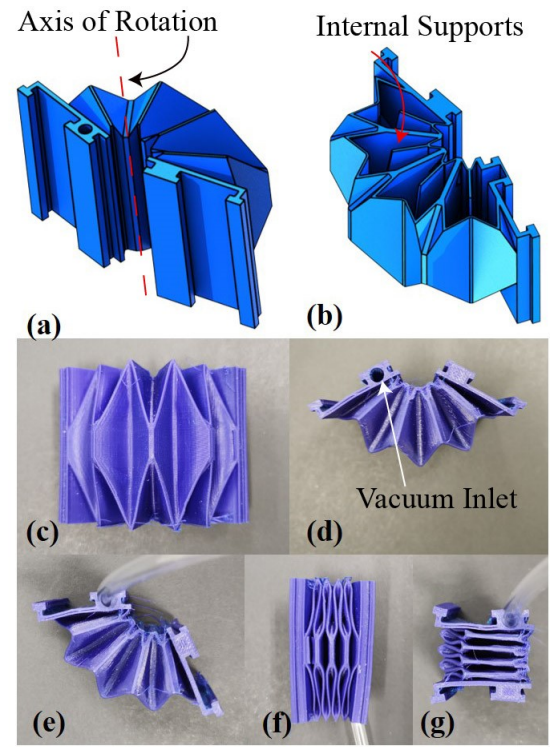

Fig. 3. (a)-(b) 3D view and cross section of vacuum actuator; (c)-(e) 3D-Printed vacuum actuator;(f)-(g) Motion and deformation of the actuator when it is vacuumized

### C. Assembly

The whole gripper is mainly composed of two parts: the body and four fingers. The principle of the design is shown in Fig. 4(a). Fig. 4(b) shows the final design of the body or rotating pentagons. White parts are made of rigid material like Polylactic acid (PLA) while the blue parts are soft material like thermoplastic elastomers (TPE). One problem implementing our design on this structure was that there was no position on the body to connect the robotic arm. We solved this by adding an extra beam inside the empty space between the four pentagons, and made screw holes on the beam to connect it to the robotic arm. In Fig. 4(c), the full assembly of the finger is shown. It has four main components: two vacuum chambers, three bio-inspired fin-ray fingertips [16], three rigid skeletons, and two flexible hinges. This kind of design combining rigid and soft materials can provide good adaptability, flexibility, and high stiffness. The fin-ray structured fingertips are for better and more stable contact with objects. Table I shows the description of different components of the assembly.

Additionally, to track the motion of the finger when actuated, two dot markers were stuck to the finger.

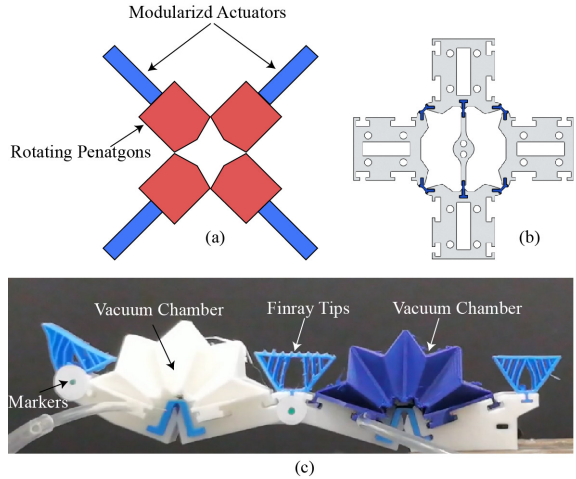

Fig. 4. (a)-(b) Design principle of the body and its final structure; (c) Design of the finger

The structure assembly is shown in Fig. 5. Fig. 5(a) and (b) shows two different stable states of the gripper whose opposite fingers are roughly at 150° to each other. The structure is locked by using eight permanent magnets. These magnets are inserted alternately into the body. Adjacent magnets have opposite magnetic poles pointing outwards. With these magnets, the structure will be locked automatically when it is folded in either way. To release the magnets and to make the body go back to a neutral state, two opposite chambers on the body will be actuated. The gripper has 12 chambers, 4 on the body and 2 on every finger, and each one can be controlled independently.

Even though strong magnets are used inside the body, it still remains stable in middle state (shown in Fig. 6(a)). The reason for this stability is that the magnets are not so powerful that they can move the weight of the fingers, but can lock when in proximity with each other. The middle state is called cross-link mode and the closed state is called parallel mode (shown in Fig. 5(a) and (b)). Fig. 6 shows how the magnets are installed.

*D. Fabrication*

One of the reasons for proposing this design is its low-cost operation. The whole structure is fabricated using a simple commercial 3D-printer. To print soft structures successfully, a printer with short-distance extruder is recommended and thus we made some custom changes to the commercial printer to realize this functionality. All vacuum actuators are printed with TPE which has 83A shore hardness, and all rigid parts are with PLA. The fin-ray tips are printed with TPU that has 95A shore hardness to provide enough stiffness and flexibility to hold objects.

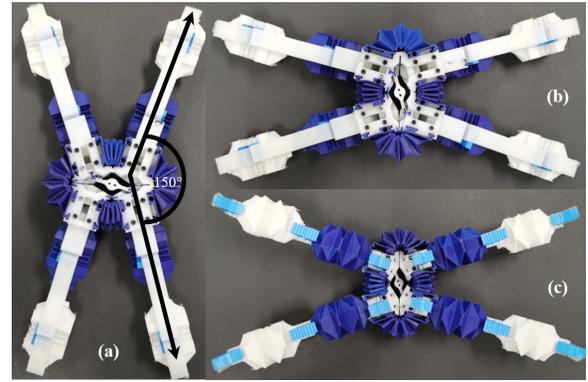

Fig. 5. (a)-(b) Two stable states of the gripper; (c) Bottom view of the gripper

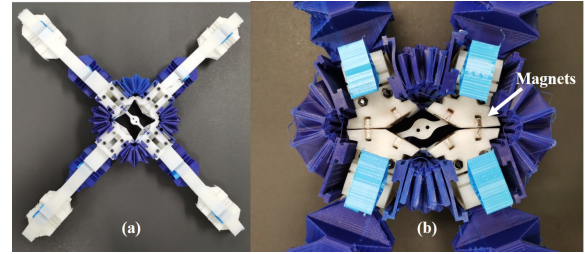

Fig. 6. (a) Middle state of the gripper, all fingers are equally spaced around the center; (b) Position and arrangement of the magnets

## III. EXPERIMENTS

In the experiment part, four major experiments were conducted:
- Repeatability and motion tracking test.
- Output force tests.
- Crawling motion test.
- Grasping tests in both modes.

Fig. 7 demonstrates how the vacuum actuator bends the finger. The whole motion from the initial state, to being fully contracted lasts for less than 1s. In the first experiment, shown in Fig. 8, three sets of tests were carried out: the first and second had both chambers actuated separately, while in the last one, both were actuated together From Fig. 8(a)-(c), when only chamber 2 is actuated, the finger can bend up to 120°. The green line is the trajectory of the tip marker in three rounds back and forth. The trajectory was generated using the software Kinovea. This is an open-source software used widely by researchers for video graphic analysis. Studies have shown that it gives results with an accuracy of 95% [17]. As only the hinges in the finger are made of flexible material, the repeatability of the finger is satisfactory. The zero correction error in several rounds is smaller than the error of the tracking algorithm and thus invisible, which means in the video we recorded the fingertip can always return to its original position once vacuum is removed. In Fig. 8(d), only chamber 1 is actuated bending the fingertip up to 105°. In Fig. 8(f), two chambers are actuated simultaneously, causing the finger skeleton to rotate 144°. This angle, coupled with 105° rotated by chamber 1, will make the fingertip to rotate more than 180°.

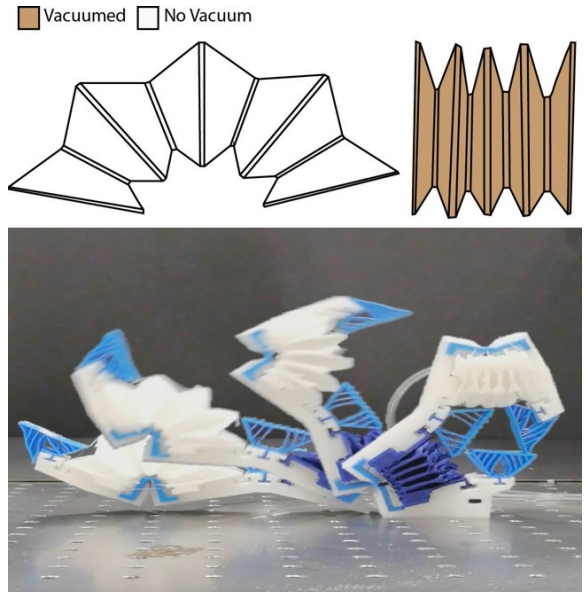

Fig. 7. Explanation of chamber motion and time-lapse photography of the finger's motion

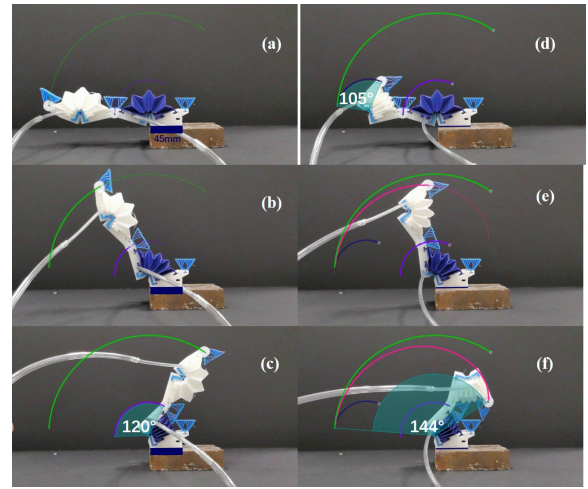

Fig. 8. (a)-(c) Trajectory of two markers when only chamber 2 is activated; (d) Trajectory of middle marker when only chamber 1 is activated; (e)-(f) Trajectory of outside marker when both chambers are activated

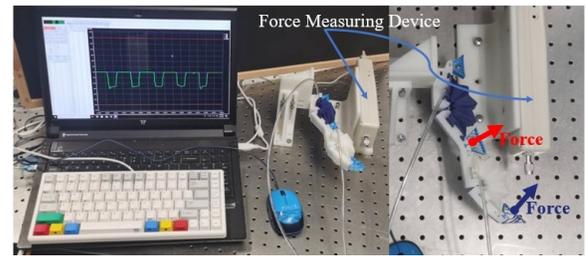

Fig. 9. Output force measurement (Force in red is the first stage output force, force in blue is the output force of fingertip)

As the movement range is sufficient, the maximum output force that the gripper can generate is important. We conducted two measuring experiments regarding the output forces in different positions. As shown in Fig. 9, we tested two output forces perpendicular to the finger. The force labeled in red is called output force in first stage, and the force in blue is called the output force of fingertip. The results are shown in Fig. 10. The maximum force that we observed in the fingertip is 15.04N, the maximum force that we recorded in the first stage is 11.13N. As the airflow created disturbances in the finger there is a possibility that these values are not very accurate. For each output force, we recorded the data of 5 repeated rounds, the force in the first stage is stable and the force in the fingertip has more deviation because the two pneumatic chambers in series will create more instability.

The third experiment was a crawling test. In this case we regard the gripper as a quadruped. In one step of movement, two adjacent fingers will be actuated, bending to the endpoint and then released. The principle which allows the quadruped to move is shown in Fig. 12. We attached four silicone pads on the top side of each fingertip. With the finger fully contracted, the silicone pad will touch the ground and produce a large friction force. After releasing the vacuum, the elastic deformation inside the soft components will recover, while the actuated fingertip will not move because of the friction force, so the whole body will be pushed forward.

Fig. 11 shows the timeline of the whole moving and turning process. Because of the asymmetry inside the whole structure, the quadruped is not moving straight forward but is a little inclined. The friction force that each fingertip can generate is also not uniform which results in an inaccurate motion. The turning structure works well, the whole robot turns through more than 90° in less than 2s, and after the direction is turned, no more vacuum is needed because of the magnetic lock.

The last experiment, and the most important, shows a promising result. In Fig. 13(a)-(b), we connect the gripper to a Universal Robotic arm to test its grasping capability. When in cross-link mode, it performs well - grasping objects that have central symmetry, like a UAV or a basketball. When handling a rectangular box that weighs 1 kg, we switch the gripper to the parallel mode and grasp the box from two sides. The gripper successfully holds the box tightly and steadily.

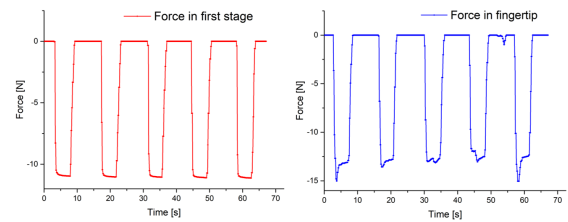

Fig. 10. Output force measurement results

TABLE I
DESCRIPTION OF DIFFERENT COMPONENTS OF THE ASSEMBLY

| Part | Components | Material | Shore Hardness | Poisson's Ratio | Usage |
|---|---|---|---|---|---|
| Vacuum Chamber | Vacuum-driven origami actuator | TPE | 83A | 0.48 to 0.50 | Actuation for whole structure |
| Rotating Pentagons | Pentagons | PLA | 75A | 0.33 | Switching between modes, |
|  | Hinges | TPE | 83A | 0.48 to 0.50 | chambers apply torque and pentagons |
|  | 4 Vacuum Chambers | TPE | 83A | 0.48 to 0.50 | bend at hinges which are soft |
| Finger | Skeleton | PLA | 75A | 0.33 | Used for gripping action and |
|  | Hinges | TPE | 83A | 0.48 to 0.50 | crawling motion of the robot |
|  | 2 Vacuum Chambers | TPE | 83A | 0.48 to 0.50 |  |
|  | Finray-tips | TPU | 95A | 0.4 |  |
| **Assembly contains one rotating pentagon and four fingers** ||||||

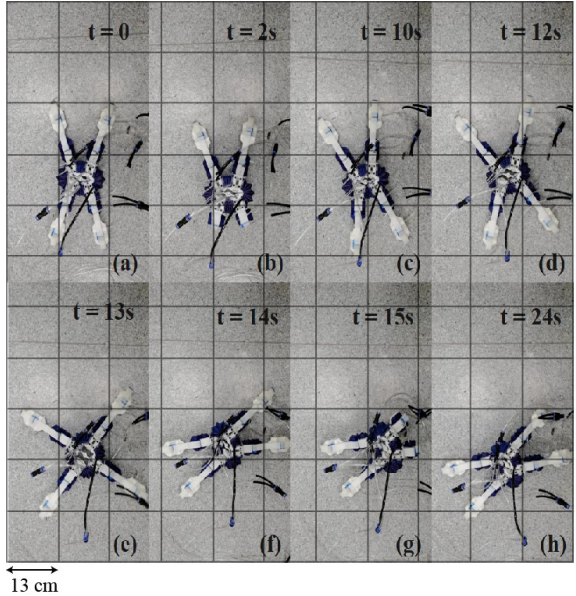

Fig. 11. (a)-(c) The quadruped is moving forwards; (d)-(f) The quadruped is turning fast by switching the body configuration; (g)-(h) The quadruped is moving to the right

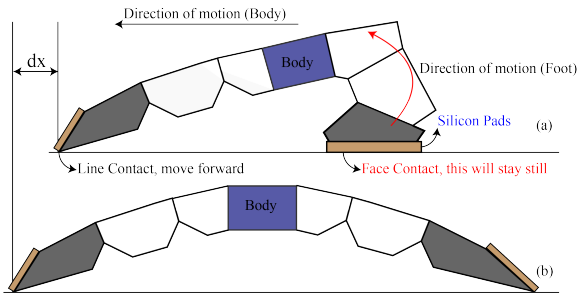

Fig. 12. Explanation of the crawling principle of the robot

To make a comparison, we conducted another test that controls the gripper to grasp the same box, but in cross-link mode. In this way, the grasping failed because the gripper is not able to put enough force on each side, and one fingertip just slides off (shown in Fig. 14). We tested this 10 times and recorded exactly the same failures.

## IV. DISCUSSION

This paper presents the design and development of an auxetic switching structure that can be used for robotic applications. To demonstrate this concept, a novel vacuum actuator is proposed. Combining this actuator with the auxetic structure, two applications are developed, a gripper and a crawling robot. Experiments showed that the range of motion of the fingertip exceeds 180°, which is impressive. The results of the gripper are also very promising. The proposed structure works well in general grasping scenarios. Although the total cost of materials is less than 20 USD, it still has good durability. However, there are still many aspects that can be improved. Initially we wanted to demonstrate that a simple auxetic structure inside the gripper can achieve passive grasping capability, meaning when the gripper is dealing with a different type of object, it will switch between parallel and cross-link mode automatically without any extra actuation. The reasons we failed to demonstrate this are twofold: 1. the folding stiffness of the body is too high when using vacuum actuation, as the counter-acting force is not enough to fold itself; 2. the fingers we made are too long, the force does not transmit effectively to the body but is dissipated because of the flexibility.

## V. ACKNOWLEDGEMENT

This research work is supported in part by the Innovation and Technology Fund of the Government of the Hong Kong Special Administrative Region (Project No. ITS/376/16FP, ITS/018/17FP, ITS/104/19FP). The authors would also like to acknowledge the help rendered by Mr. Zicheng Kan (Graduate Student, HKUST) and Mr. Yazhan Zhang (Graduate Student, HKUST) during this study.

## REFERENCES

[1] J. N. Grima and K. E. Evans, "Auxetic behavior from rotating squares," 2000.
[2] K. E. Evans, "Auxetic polymers: a new range of materials," *Endeavour*, vol. 15, no. 4, pp. 170–174, 1991.
[3] W. Yang, Z.-M. Li, W. Shi, B.-H. Xie, and M.-B. Yang, "Review on auxetic materials," *Journal of materials science*, vol. 39, no. 10, pp. 3269–3279, 2004.
[4] T.-C. Lim, *Auxetic materials and structures*. Springer, 2015.
[5] J. N. Grima, A. Alderson, and K. Evans, "Auxetic behaviour from rotating rigid units," *Physica status solidi (b)*, vol. 242, no. 3, pp. 561–575, 2005.
[6] M. Manti, T. Hassan, G. Passetti, N. D'Elia, C. Laschi, and M. Cianchetti, "A bioinspired soft robotic gripper for adaptable and effective grasping," *Soft Robotics*, vol. 2, no. 3, pp. 107–116, 2015.

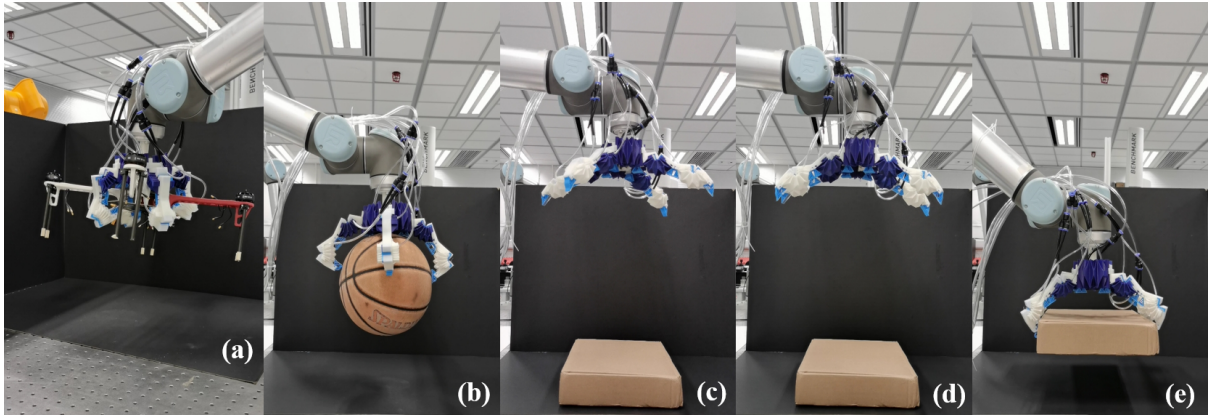

Fig. 13. (a) Grasping test of a UAV using cross-link mode; (b) Grasping test of a basketball using equally-spaced mode; (c)-(e) Grasping test of a rectangular box using parallel mode, from (c) to (d), the gripper changed working mode

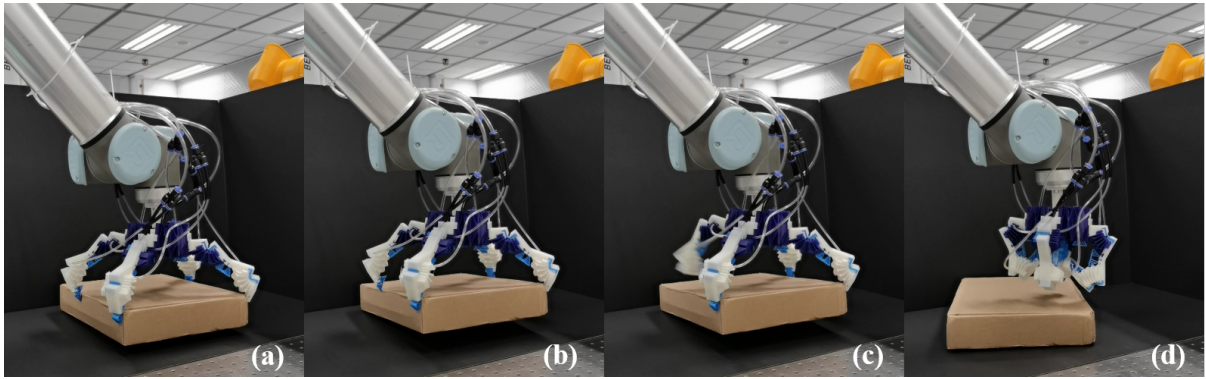

Fig. 14. (a)-(d) Grasping test of the same rectangular box but using cross-link mode and failed.

[7] L. A. Al Abeach, S. Nefti-Meziani, and S. Davis, "Design of a variable stiffness soft dexterous gripper," *Soft robotics*, vol. 4, no. 3, pp. 274–284, 2017.

[8] A. M. Abdullah, X. Li, P. V. Braun, J. A. Rogers, and K. J. Hsia, "Self-folded gripper-like architectures from stimuli-responsive bilayers," *Advanced Materials*, vol. 30, no. 31, p. 1801669, 2018.

[9] R. L. Truby, M. Wehner, A. K. Grosskopf, D. M. Vogt, S. G. Uzel, R. J. Wood, and J. A. Lewis, "Soft somatosensitive actuators via embedded 3d printing," *Advanced Materials*, vol. 30, no. 15, p. 1706383, 2018.

[10] Y. Tang, Y. Li, Y. Hong, S. Yang, and J. Yin, "Programmable active kirigami metasheets with more freedom of actuation," *Proceedings of the National Academy of Sciences*, vol. 116, no. 52, pp. 26 407–26 413, 2019.

[11] D. Rus and M. T. Tolley, "Design, fabrication and control of origami robots," *Nature Reviews Materials*, vol. 3, no. 6, p. 101, 2018.

[12] R. Xie, M. Su, Y. Zhang, M. Li, H. Zhu, and Y. Guan, "Pisrob: A pneumatic soft robot for locomoting like an inchworm," in *2018 IEEE International Conference on Robotics and Automation (ICRA)*. IEEE, 2018, pp. 3448–3453.

[13] L. Qin, X. Liang, H. Huang, C. K. Chui, R. C.-H. Yeow, and J. Zhu, "A versatile soft crawling robot with rapid locomotion," *Soft robotics*, vol. 6, no. 4, pp. 455–467, 2019.

[14] C. D. Onal, R. J. Wood, and D. Rus, "An origami-inspired approach to worm robots," *IEEE/ASME Transactions on Mechatronics*, vol. 18, no. 2, pp. 430–438, 2012.

[15] A. R. Deshpande, Z. T. H. Tse, and H. Ren, "Origami-inspired bi-directional soft pneumatic actuator with integrated variable stiffness mechanism," in *2017 18th International Conference on Advanced Robotics (ICAR)*. IEEE, 2017, pp. 417–421.

[16] W. Crooks, G. Vukasin, M. O'Sullivan, W. Messner, and C. Rogers, "Fin ray® effect inspired soft robotic gripper: From the robosoft grand challenge toward optimization," *Frontiers in Robotics and AI*, vol. 3, p. 70, 2016.

[17] A. Puig-Diví, C. Escalona-Marfil, J. M. Padullés-Riu, A. Busquets, X. Padullés-Chando, and D. Marcos-Ruiz, "Validity and reliability of the kinovea program in obtaining angles and distances using coordinates in 4 perspectives," *PloS one*, vol. 14, no. 6, p. e0216448, 2019.